\documentclass[lettersize,journal]{IEEEtran}
\usepackage{amsmath,amsfonts}
\usepackage{algorithmic}
\usepackage{algorithm}
\usepackage{array}
\usepackage[caption=false,font=normalsize,labelfont=sf,textfont=sf]{subfig}
\usepackage{textcomp}
\usepackage{stfloats}
\usepackage{url}
\usepackage{verbatim}
\usepackage{graphicx}
\usepackage{booktabs}
\usepackage[utf8]{inputenc}
\usepackage{threeparttable}
\usepackage{float}
\usepackage{pdflscape}
\usepackage{adjustbox}
\usepackage{cite}
\usepackage{multicol}
\usepackage{booktabs} 
\usepackage{xcolor}
\usepackage[inkscapelatex=false]{svg}

\usepackage{tabularx}
\usepackage{newunicodechar}
\newunicodechar{∼}{\textasciitilde}
\begin{document}


\newcommand{\overviewFigure}{%
\begin{figure}[!t]
\centering
\includegraphics[width=\columnwidth]{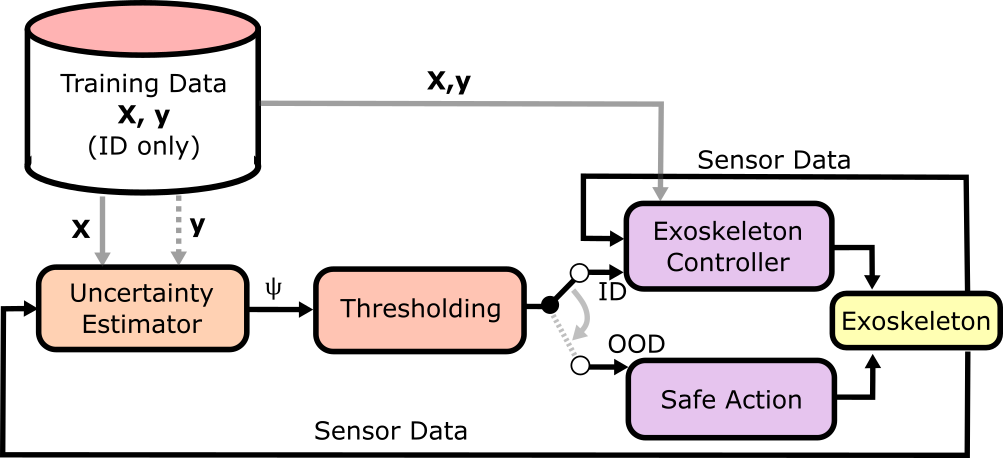}
    \caption{Overview of the uncertainty classifier system for exoskeleton control. A base exoskeleton controller is trained on sensor data (X) and a label (y), such as gait phase or biological joint torque. An ideal uncertainty classifier is trained using only in-distribution (ID) sensor data. The uncertainty classifier determines if incoming sensor data from the exoskeleton is similar to or different from the training data. When sensor data is similar to training data, the exoskeleton applies assistive torque; when sensor data differs substantially from training data, the exoskeleton switches to a fallback action intended to be safer, such as disengaging / providing no assistance.}
\label{fig:overview}
\vspace{-1em}
\end{figure}
}

\newcommand{\architectureFigure}{%
\begin{figure}[!t]
\centering
\includegraphics[width=0.7\columnwidth]{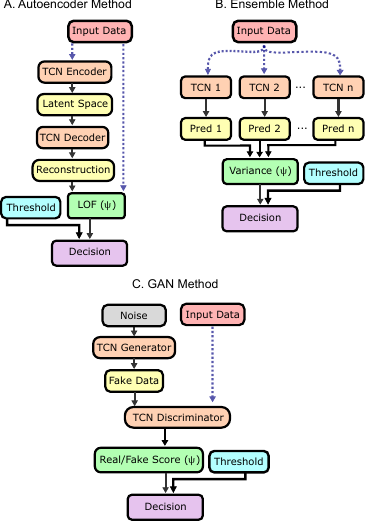}
\caption{Neural network architectures for uncertainty classification: (A) Autoencoder approach with TCN-based encoder and decoder to measure Local Outlier Factor, (B) Ensemble method using multiple TCN models to measure prediction variance,  and (C) GAN-based approach with generator and discriminator networks.}
\label{fig:architecture}
\vspace{-1em}
\end{figure}
}

\newcommand{\methodsFigure}{%
\begin{figure}[!t]
\centering
\includegraphics[width=\columnwidth]{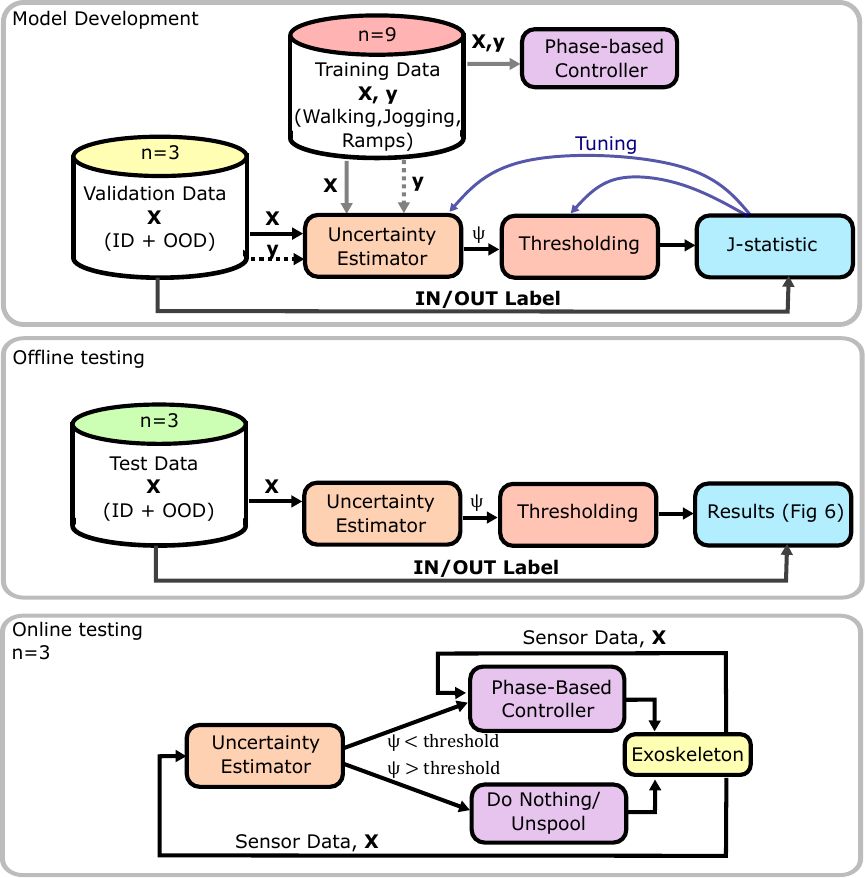}
    \caption{Overview of the methods used to develop and test the uncertainty estimators for exoskeleton control. The uncertainty estimators are trained using \train, with all models except for the gait phase ensemble training only on sensor data, while the gait phase ensemble also used gait phase labels from this dataset. Model development / tuning was based on performance on \val.   Offline testing was performed on \testOffline, which included new ID and OOD tasks and new subjects (results in Section \ref{sec:offline-results}). For online testing (\testOnline), we implemented the uncertainty estimator and phase-based controller with the ankle exoskeleton actuating and responding to the models in real time (results in Section \ref{sec:online-model-testing}.)}
\label{fig:methods}
\vspace{-1em}
\end{figure}
}

\newcommand{\offlineResultsFigure}{%
\begin{figure*}[!t]
\centering
\includegraphics[width=\textwidth]{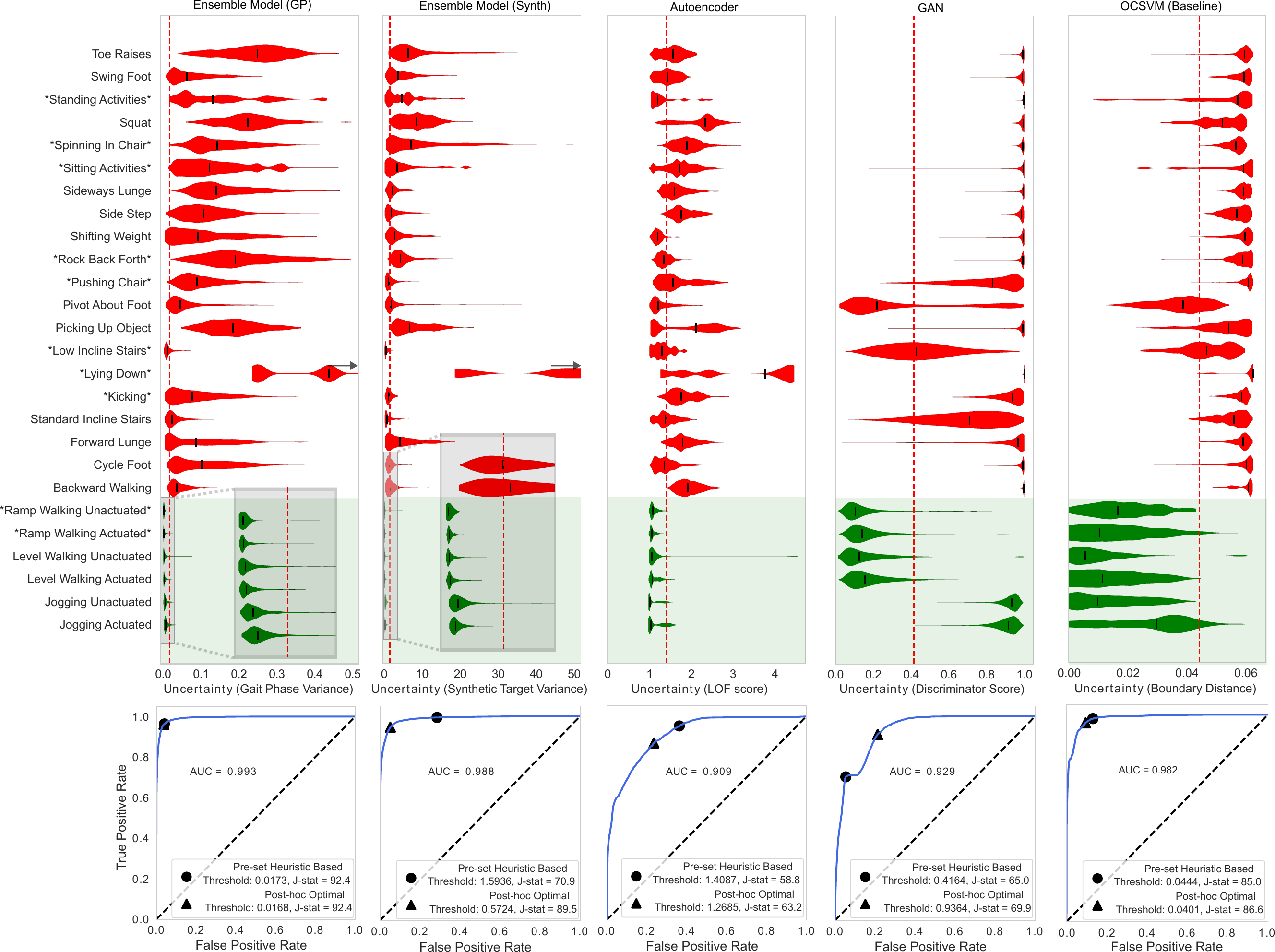}
\caption{Offline test results: Distribution of uncertainty scores across the full duration of in-distribution (green) trials and out-of-distribution (red) trials, shown on all model architectures. New tasks and tasks with substantial changes from \val to \testOffline are denoted with *stars*. The bottom row is of each model's ROC curve with the pre-set heuristic based threshold denoted by a circle, and the threshold optimized for J-statistic on \testOffline denoted by a triangle. J-statistics are computed on pooled scores; Table I reports the per-subject mean.  }
\label{fig:offline_results}
\vspace{-1em}
\end{figure*}
}

\newcommand{\sensorWindowsFigure}{%
\begin{figure*}[!t]
\centering
\includegraphics[width=0.95\textwidth]{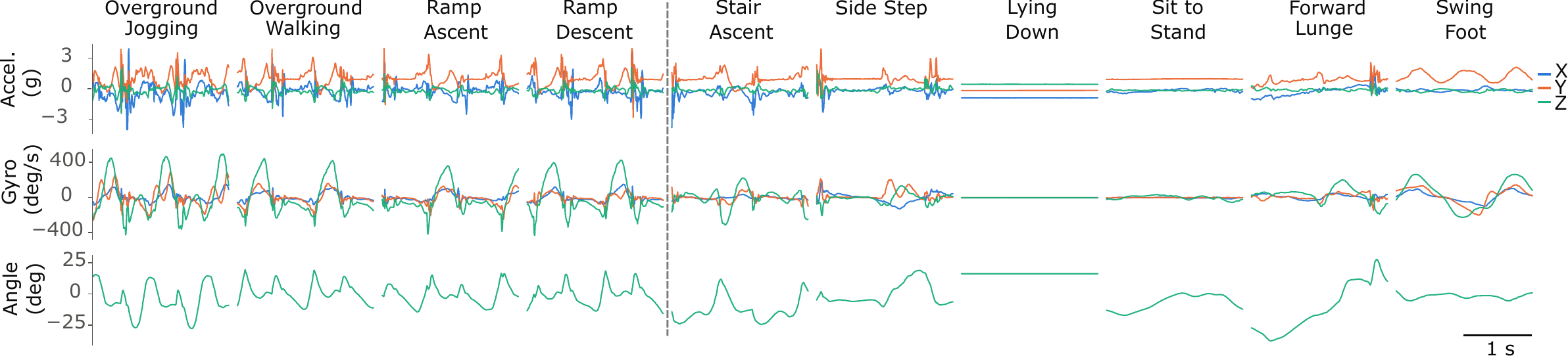}
\caption{Sensor data from the right exoboot of a subject in \testOffline, with in-distribution tasks to the left of the dotted line and out-of-distribution tasks to the right. Each task is shown with 2 seconds of sensor data, which includes IMU accelerometer data (first row), IMU gyroscope data (second row), and encoder data (third row). Windows of this sensor data are used as input for the uncertainty estimator to make its prediction.}
\label{fig:sensor_windows}
\vspace{-1em}
\end{figure*}
}

\newcommand{\onlineFigure}{%
\begin{figure*}[!t]
\centering
\includegraphics[width=0.95\textwidth]{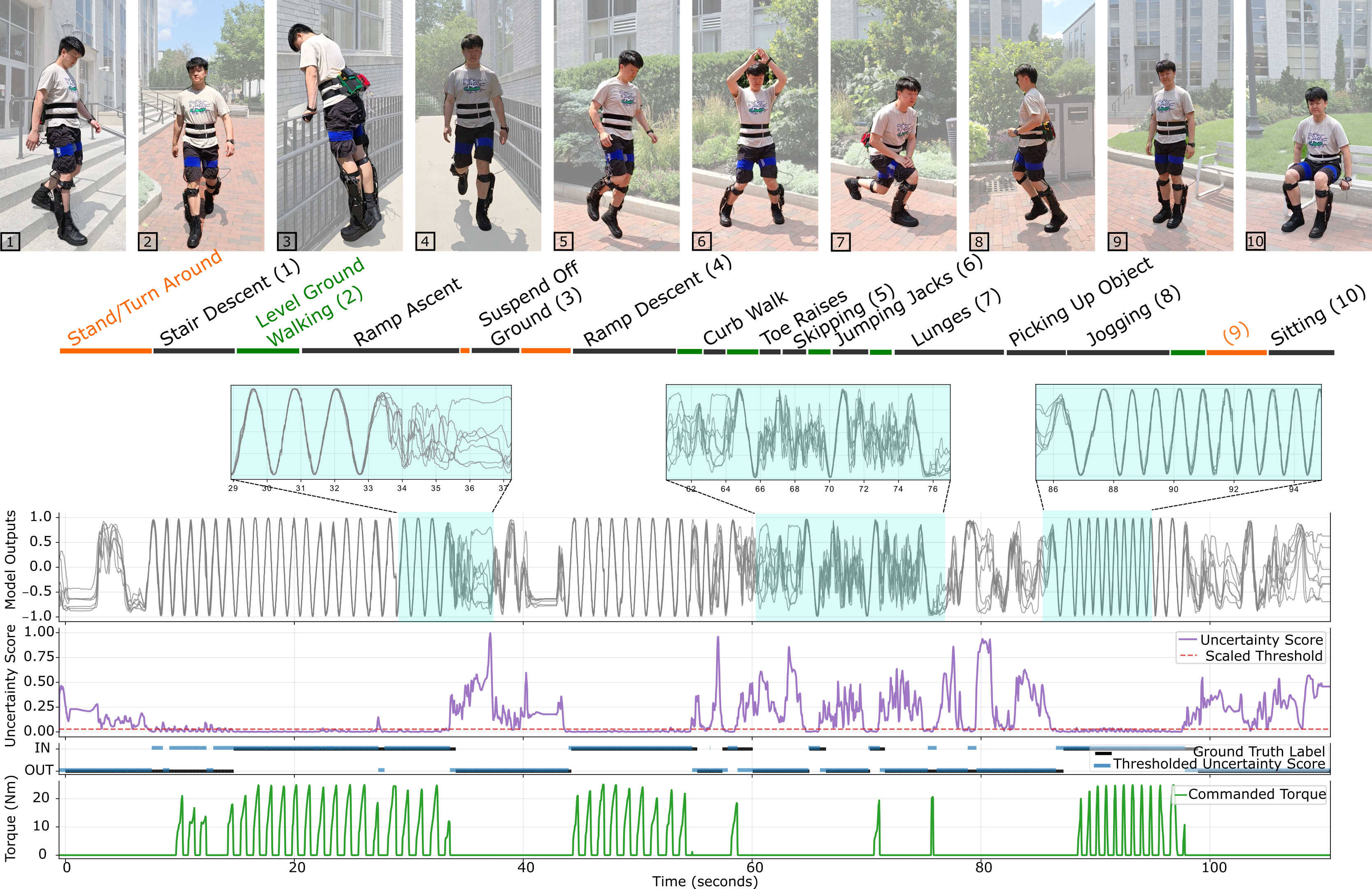}
    \caption{Participant performing various in- and out-of-distribution actions on the outdoor course, with labeled actions and corresponding anomaly score plotted below. Variability in the output of the 7 gait phase models increases during out-of-distribution actions. The effect on the user is shown through the applied torque plot (bottom row), which shows that torque is only applied when the model labels an action as in-distribution.The torque shown represents the torque applied by the left exoboot.}
\label{fig:onlineFigure}
\vspace{-1em}
\end{figure*}
}

\newcommand{\CourseMapFigure}{%
\begin{figure}[!t]
\centering
\includegraphics[width=\columnwidth]{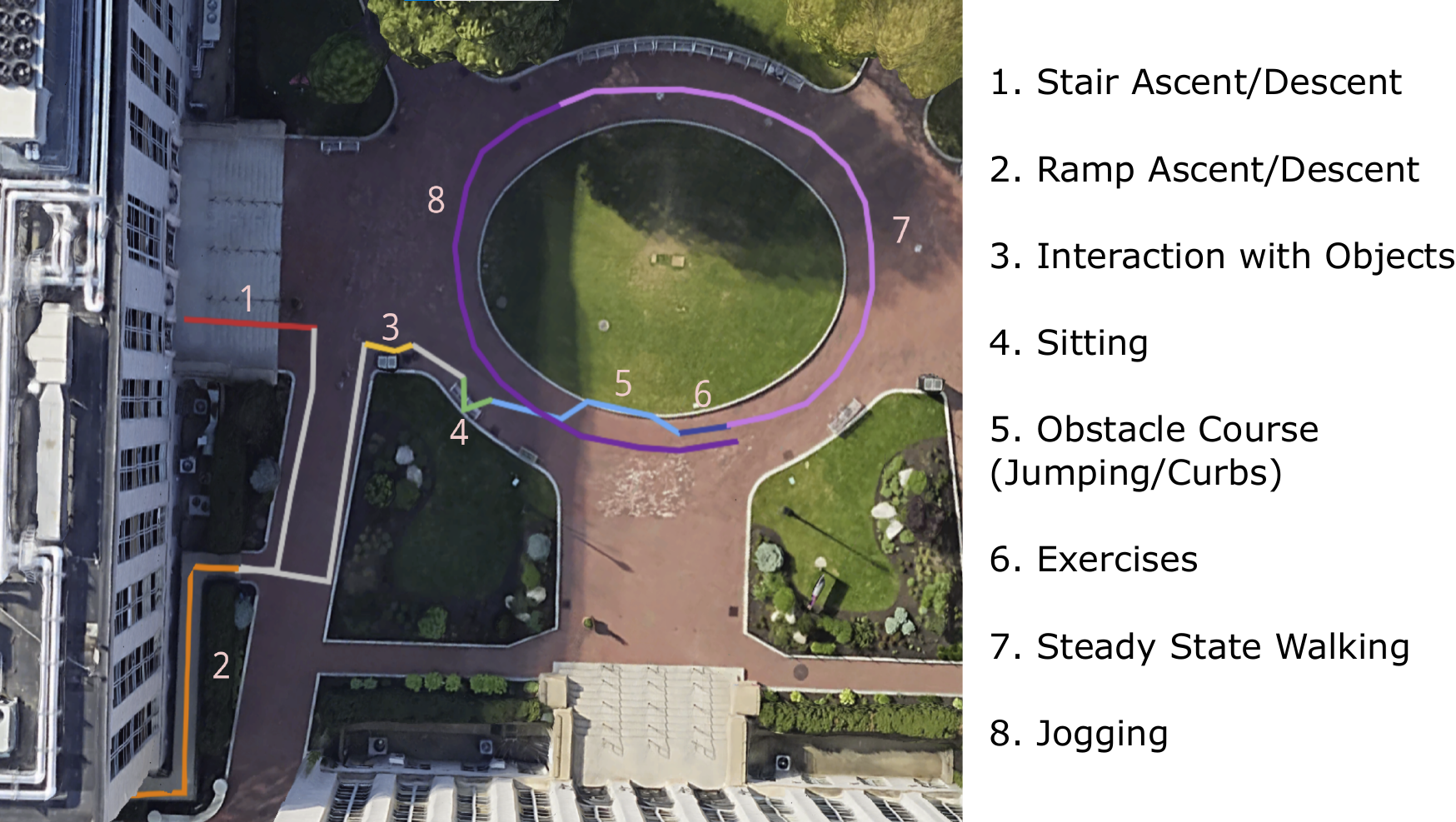}
\caption{Test route around Northeastern University's Krentzman Quadrangle: The outdoor course was designed to validate the system's performance in real-world conditions, featuring various terrains and actions. Key stations include: (1) \textbf{Stair station} (ascending/descending stairs); (2) \textbf{Ramps and suspension station} (ramp navigation and hanging); (3) \textbf{Object interaction station} (trashcan operation and object retrieval); (4) \textbf{Sitting station} (sit-stand transitions and seated movements); (5) \textbf{Obstacle course station} (curb navigation and jumping); (6) \textbf{Exercise station} (squats, lunges, and toe raises); (7) \textbf{Steady state walking} (walking at a brisk pace); (8) \textbf{Jogging} (jogging at a slow pace). }
\label{fig:course_map}
\vspace{-2em}
\end{figure}
}

\newcommand{\AutoencoderFigure}{%
\begin{figure*}[!t]
\centering
\includegraphics[width=0.9\textwidth]{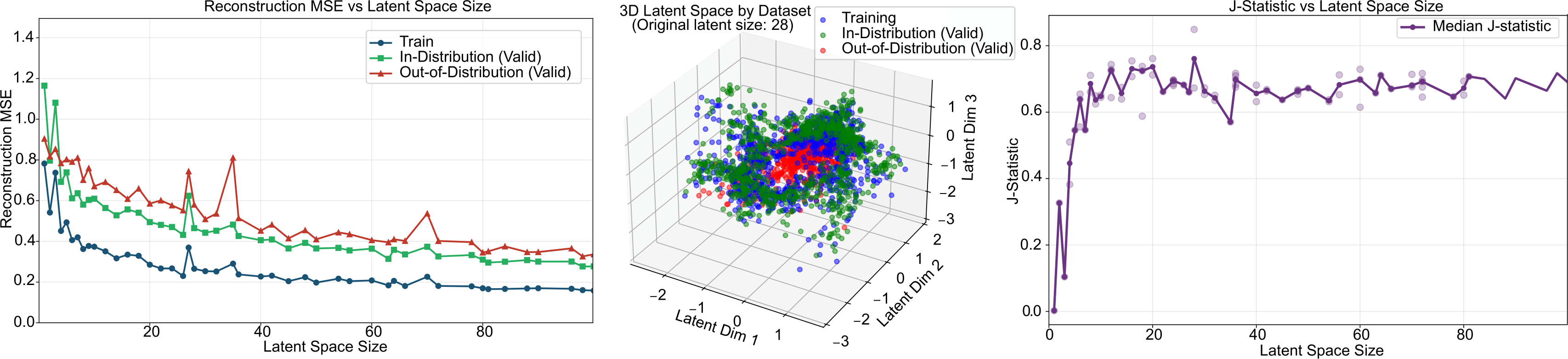}
\caption{From left to right, a) Latent Space Size vs. Reconstruction MSE of Training, In Distribution Validation, and Out of Distribution Dataset. b) 3D plot of first 3 dimensions of latent space of best performing model. c) Latent space size vs J-statistic }
\label{fig:autoencoder}
\end{figure*}
}


\newcommand{\offlineResultsTable}{%
\begin{table*}[!b]
\renewcommand{\arraystretch}{1.3}
\caption{Offline Evaluation Results Comparison of Different Architectures}
\label{tab:offline_results}
\centering
\begin{tabular}{|l|c|c|c|c|c|c|c|}
\hline
\textbf{Method} & \textbf{Accuracy (\%)} & \textbf{Precision (\%)} & \textbf{Recall (\%)} & \textbf{F1-Score (\%)} & \textbf{J-Statistic (\%)} & \textbf{ECE} & \textbf{Brier} \\
\hline
Baseline (One-Class SVM) & 89.1 $\pm$ 4.5 & 65.2 $\pm$ 11.3 & 98.2 $\pm$ 1.7 & 77.9 $\pm$ 8.2 & 85.1 $\pm$ 4.5 & 0.15 & 0.09 \\
\hline
Ensemble Method (Gait Phase) & \textbf{96.1 $\pm$ 1.3} & \textbf{85.2 $\pm$ 5.2} & 96.2 $\pm$ 2.8 & \textbf{90.3 $\pm$ 3.8} & \textbf{92.3 $\pm$ 3.5} & \textbf{0.03} & \textbf{0.03} \\
\hline
Ensemble Method (Synthetic Target) & 76.9 $\pm$ 2.6 & 45.7 $\pm$ 4.7 & \textbf{99.4 $\pm$ 0.8} & 62.5 $\pm$ 4.6 & 70.9 $\pm$ 3.3 & 0.22 & 0.13 \\
\hline
Autoencoder & 69.8 $\pm$ 10.9 & 40.0 $\pm$ 12.2 & 94.7 $\pm$ 6.1 & 55.7 $\pm$ 12.8 & 58.5 $\pm$ 16.9 & 0.26 & 0.18 \\
\hline
GAN & 89.4 $\pm$ 2.7 & 74.9 $\pm$ 13.0 & 68.7 $\pm$ 3.8 & 71.5 $\pm$ 7.8 & 63.0 $\pm$ 6.5 & 0.06 & 0.09 \\
\hline
\end{tabular}
\\[5pt]
{\small Values are reported as mean $\pm$ standard deviation across three subjects in the offline test set.}
\end{table*}
}

\newcommand{\onlineOverallResultsTable}{%
\begin{table*}[!b]
\renewcommand{\arraystretch}{1.3}
\caption{Transition vs. Steady State Performance and Overall Results}
\label{tab:ensemble_transitions_overall}
\centering
\begin{tabular}{|l|c|c|c|c|c|}
\hline
\textbf{Dataset}& \textbf{Accuracy (\%)}& \textbf{Precision (\%)}& \textbf{Recall (\%)}& \textbf{F1-Score}& \textbf{J-Statistic}\\
\hline
\multicolumn{6}{|c|}{\textbf{Transitions vs Steady State}} \\
\hline
Transition (entire dataset)  & 65.5 $\pm$ 7.8 & 69.4 $\pm$ 5.4 & 49.4 $\pm$ 21.6 & 55.8 $\pm$ 17.0 & 30.1 $\pm$ 15.6 \\
\hline
Steady State (entire dataset)& 89.1 $\pm$ 1.8 & 89.4 $\pm$ 3.6 & 92.3 $\pm$ 2.1 & 90.8 $\pm$ 1.2 & 77.0 $\pm$ 4.7 \\
\hline
Transition (No stair section) & 62.7 $\pm$ 8.9 & 71.4 $\pm$ 6.6 & 48.1 $\pm$ 24.4 & 54.4 $\pm$ 19.8 & 26.6 $\pm$ 16.4 \\
\hline
Steady State (No stair section)& 92.4 $\pm$ 1.8 & 96.3 $\pm$ 1.0 & 92.3 $\pm$ 2.1 & 94.2 $\pm$ 1.2 & 84.7 $\pm$ 3.8 \\
\hline
\multicolumn{6}{|c|}{\textbf{Overall}} \\
\hline
All Actions Total & 87.6 $\pm$ 1.6 & 88.6 $\pm$ 3.1 & 90.1 $\pm$ 2.6 & 89.3 $\pm$ 1.0 & 74.4 $\pm$ 4.0 \\
\hline
All Actions Except Stairs Total & 90.4 $\pm$ 1.9 & 95.2 $\pm$ 0.3 & 90.1 $\pm$ 2.6 & 92.5 $\pm$ 1.4 & 81.0 $\pm$ 3.3 \\
\hline
\end{tabular}
\\[5pt]
{\small Values are reported as mean $\pm$ standard deviation across three subjects in the outdoor test set.}
\end{table*}
}

\newcommand{\onlineStationsResultsTable}{%
\begin{table}[!b]
\renewcommand{\arraystretch}{1.3}
\caption{Detailed Results for Ensemble Gait Phase Method by Action Group}
\label{tab:ensemble_results_by_station}
\centering
\begin{tabular}{|l|c|}
\hline
\textbf{Section/Action Group} & \textbf{Accuracy (\%)}\\
\hline
Level Ground Walking & 89.4 $\pm$ 3.1 \\
\hline
Incline and Decline Walking & 96.7 $\pm$ 2.9 \\
\hline
Jogging & 86.5 $\pm$ 6.0 \\
\hline
Stairs & 71.1 $\pm$ 13.3 \\
\hline
Slow Out of Distribution Tasks & 90.7 $\pm$ 1.6 \\
\hline
Fast Out of Distribution Tasks & 86.7 $\pm$ 5.4 \\
\hline
\end{tabular}
\\[5pt]
{\small Values are reported as mean $\pm$ standard deviation across three subjects in the outdoor test set.}
\vspace{-2em}
\end{table}
}


\newcommand{\noveltyTable}{%
\begin{table}[H]
\renewcommand{\arraystretch}{1.2}
\setlength{\tabcolsep}{4pt}
\caption{\small Subject and Activity Novelty Across Study Stages}
\label{tab:novelty_overview}
\centering
\scriptsize
\begin{tabularx}{\columnwidth}{|X|c|c|c|c|}
\hline
\textbf{Item / Activity category} & {\train} & {\val} & {\testOffline} & {\testOnline} \\
\hline
\multicolumn{5}{|c|}{\textbf{Participants}} \\
\hline
Number of subjects & 9 & 3 & 3 & 3 \\
\hline
\multicolumn{5}{|c|}{\textbf{In-Distribution (cyclic; actuated \& unactuated)}} \\
\hline
Level-ground / treadmill walking (multi-speed) & $\bullet$ & $\bullet$ & $\bullet$ & $\bullet$ \\
\hline
Jogging & $\bullet$ & $\bullet$ & $\bullet$ & $\bullet$ \\
\hline
Ramp / incline walking & $\bullet$ & -- & $\bullet$* & $\bullet$ \\
\hline
\multicolumn{5}{|c|}{\textbf{Out-of-Distribution}} \\
\hline
Backward walking & -- & $\bullet$ & $\bullet$ & -- \\
\hline
Turning / pivoting in place & -- & $\bullet$ & $\bullet$ & $\bullet$ \\
\hline
Side-stepping & -- & $\bullet$ & $\bullet$ & -- \\
\hline
Skipping & -- & $\bullet$ & $\bullet$ & $\bullet$ \\
\hline
Jumping / hopping (incl.\ jumping jacks) & -- & $\bullet$ & -- & $\bullet$ \\
\hline
Standard (indoor) stairs & -- & $\bullet$ & $\bullet$ & $\bullet$ \\
\hline
Low-incline / outdoor stairs & -- & -- & $\bullet$* & $\bullet$ \\
\hline
Curb navigation (on/off curb) & -- & -- & -- & $\bullet$ \\
\hline
Non-weight-bearing foot motion (cycling / swinging one foot) & -- & $\bullet$ & $\bullet$ & -- \\
\hline
Standing strength exercises (lunges, squats, toe raises) & -- & $\bullet$ & $\bullet$ & $\bullet$ \\
\hline
Standing / weight shifting / leaning & -- & $\bullet$ & $\bullet$* & $\bullet$ \\
\hline
Seated \& chair-based (sit-to-stand, tapping, spinning, dangling) & -- & $\bullet$ & $\bullet$* & $\bullet$ \\
\hline
Lying down & -- & -- & $\bullet$* & -- \\
\hline
Kicking & -- & -- & $\bullet$* & -- \\
\hline
Pushing a rolling chair & -- & -- & $\bullet$* & -- \\
\hline
Object interaction (picking up / handling objects) & -- & $\bullet$ & $\bullet$ & $\bullet$* \\
\hline
Suspending feet off ground (hanging) & -- & -- & -- & $\bullet$* \\
\hline
\end{tabularx}
\\[4pt]
{\footnotesize $\bullet$ = task category included at this stage; $\bullet$* = included and newly added or substantially expanded relative to the previous stage (starred tasks in Fig.~\ref{fig:offline_results} for the offline test set); -- = not included. Columns denote the training (\train), validation (\val), offline test (\testOffline), and online test (\testOnline) datasets. Offline and online test subjects are entirely novel; only the validation set shares one subject with training.}
\end{table}
}


\newcommand{\allmodelspecstableonecol}{%
\begin{table}[!t]
\centering
\scriptsize
\renewcommand{\arraystretch}{1.0}
\caption{Complete Specifications for All Anomaly Detection Models}
\label{tab:all_model_specs}
\begin{tabularx}{\columnwidth}{|l|X|}
\hline
\textbf{Parameter} & \textbf{Value/Description} \\
\hline
\multicolumn{2}{|c|}{\textbf{Common Data Preparation (All Models)}} \\
\hline
Subjects & S02-S10 (9 subjects) \\
Trials & T01, T02, T04-T12 (11 trials) \\
Window size & 175 samples (1 second at 175 Hz) \\
Input channels & 16 (bilateral sensor data) \\
Input features & (accel\_[x,y,z], gyro\_[x,y,z], ankle\_angle, ankle\_velocity) $\times$ 2 \\
Scaling & StandardScaler (pre-trained) \\
Smoothing & SMA (88 samples, 0.5 second window) \\
\hline
\multicolumn{2}{|c|}{\textbf{Ensemble Models}} \\
\hline
Architecture & Multi-branch convolutional predictor, 7 branches, 3 layers/branch \\
Filters/Kernel & 30 filters/layer, kernel size 20 \\
Activation/Norm & ReLU, BatchNorm1d after each convolution \\
Training params & Step: 10, Batch: 1024, LR: 0.001, Optimizer: Adam \\
Loss function & MSE \\
Early stopping & 10 epochs patience during LOSO; final model uses LOSO-determined epochs \\
Anomaly scoring & Variance across 7 branch predictions, 99.5th percentile threshold \\
\hline
\multicolumn{2}{|c|}{\textbf{Ensemble: Gait Phase Model}} \\
\hline
Target & Physiological gait phase signals, dual outputs (left/right) \\
Output/Loss & Tanh activation (range $[-1, 1]$), MSE loss, Final epochs: 13$^1$ \\
\hline
\multicolumn{2}{|c|}{\textbf{Ensemble: Synthetic Target Model}} \\
\hline
Target & Correlation-based: $\sum_{i>j} \text{corr}(x_i, x_j)$, single output \\
Output/Loss & Linear activation, MSE loss, Final epochs: 34$^1$ \\
\hline
\multicolumn{2}{|c|}{\textbf{Autoencoder Model}} \\
\hline
Architecture & Conv. autoencoder with temporal compression \\
Encoder/Decoder & Conv1d(16→19→24→bottleneck), kernel sizes: 15, 17, 19; Mirror decoder \\
Compression & Time dimension of latent space: 7, Filter dimension of latent space: 4, Latent space size: 28 \\
Activation/Norm & ReLU, BatchNorm1d after each layer, No dropout \\
Training params & Step: 10, Batch: 1024, LR: 0.001*(1024/16), Optimizer: Adam \\
Loss function & MSE \\
Early stopping & 80/20 train/valid split, patience=5 \\
Anomaly scoring & LOF(latent space), 99.5th percentile threshold \\
\hline
\multicolumn{2}{|c|}{\textbf{GAN Model}} \\
\hline
Generator & ConvTranspose1d(10→10→12→16), kernel: 3, LeakyReLU(0.2) \\
Discriminator & Conv1d(16→30→30) w/ spectral norm, kernel: 5, ReLU, dropout: 0.2 \\
Output/Latent & Sigmoid (discriminator), Latent: 10 channels $\times$ 35 time steps \\
Training params & Step: 20, Batch: 256, LR: G=2e-4, D=5e-5, Exp. decay ($\gamma$=0.99) \\
Optimizer/Loss & Adam ($\beta$=(0.5, 0.999)), Binary Cross-Entropy \\
Training & 500 epochs (fixed), 80/20 split, 5 G updates per 1 D update \\
Checkpointing & Every 5 epochs \\
Anomaly scoring & Discriminator output 1- D(x), 99.5th percentile threshold \\
\hline
\multicolumn{2}{|c|}{\textbf{Baseline: One-Class SVM}} \\
\hline
Feature extraction & Flatten (time\_stride=5): 35 timesteps $\times$ 16 channels = 560 features, amplitude-normalized (÷ window RMS) \\
Dimensionality reduction & PCA to 20 components, whitened (unit variance) \\
Detector & One-class SVM, RBF kernel \\
Hyperparameters & $\nu$=0.01, $\gamma$=0.50 (10$\times$ scikit-learn `scale') \\
Anomaly scoring & Negative signed distance to decision boundary, 99.5th percentile threshold \\
\hline
\multicolumn{2}{|c|}{\textbf{Training Strategy Comparison}} \\
\hline
Ensemble & Stage 1: LOSO validation, Stage 2: Train on all subjects \\
AE/GAN & Single stage: Train on all subjects with validation split \\
Baseline (OCSVM) & Two-stage sweep: Stage 1 representation/kernel selection, Stage 2 refit + threshold on all subjects \\
\hline
\end{tabularx}
\end{table}
}

\title{Uncertainty-Aware Ankle Exoskeleton Control}
\author{
Fatima Mumtaza Tourk\textsuperscript{*},~\IEEEmembership{Student Member,~IEEE,}
\and
Bishoy Galoaa\textsuperscript{*},~\IEEEmembership{Student Member,~IEEE,}
\and
Sanat Shajan\textsuperscript{*},
\and
Aaron J. Young,~\IEEEmembership{Senior Member,~IEEE,}
\and
Michael Everett,~\IEEEmembership{Member,~IEEE,}
\and
Max K. Shepherd,~\IEEEmembership{Member,~IEEE}
\thanks{\textsuperscript{*}These authors contributed equally to this work.}
\thanks{This work was supported in part by U.S. National Science Foundation FRR Award under Grant 2328050 and Grant 2328051 and in part by Northeastern University's Research Computing team. (Corresponding author: Fatima Mumtaza Tourk.) This work involved human subjects or animals in its research. Approval of all ethical and experimental procedures and protocols was granted by Northeastern Institutional Review Board, \#22-11-29, and performed in line with the Declaration of Helsinki. 
\par
F. M. Tourk is with the Department of Mechanical Engineering, Northeastern University, Boston, MA 02115 USA (e-mail: tourk.f@northeastern.edu).
\par
B. Galoaa is with the College of Engineering, Northeastern University, Boston, MA 02115 USA.
\par
S. Shajan is with the Khoury College of Computer Sciences, Northeastern University, Boston, MA 02115 USA.
\par
M. Everett is with the Department of Electrical \& Computer Engineering, Khoury College of Computer Sciences, and the Institute for Experiential Robotics, Northeastern University, Boston, MA 02115 USA.
\par
A. J. Young is with the Woodruff School of Mechanical Engineering and the Institute for Robotics and Intelligent Machines, Georgia Institute of Technology, Atlanta, GA 30332-0405 USA.
\par
M. K. Shepherd is with the Department of Mechanical Engineering, the Department of Physical Therapy, Movement, and Rehabilitation Science, and the Institute for Experiential Robotics, Northeastern University, Boston, MA 02115 USA.
}
}
\IEEEaftertitletext{\vspace{-1.5\baselineskip}}
\maketitle

\footnotetext{This work was supported by U.S. National Science Foundation FRR Award under Grant 2328050 and Grant 2328051 and by Northeastern University's Research Computing team. This work involved human subjects or animals in its research. Approval of all ethical and experimental procedures and protocols was granted by Northeastern Institutional Review Board, \#22-11-29, and performed in line with the Declaration of Helsinki.}

\begin{abstract}
Lower limb exoskeletons show promise to assist human movement, but their utility is limited by controllers designed for discrete, predefined actions in controlled environments, restricting their real-world applicability. We present an uncertainty-aware control framework that enables ankle exoskeletons to operate across diverse scenarios by automatically disengaging when encountering unfamiliar movements. Our approach uses an uncertainty estimator that takes onboard exoskeleton sensor (IMU and encoder) data to classify movements as similar (in-distribution) or different (out-of-distribution) relative to actions in the training set. We evaluated three deep architectures (model ensembles, autoencoders, and generative adversarial networks) against a one-class SVM baseline on an offline dataset, with an ensemble of gait-phase estimators performing best (F1=90.3\%). An online test demonstrated the ability of our uncertainty estimator to turn assistance on and off as the user transitioned between in-distribution and out-of-distribution tasks F1=89.3\%). This new framework provides a path toward exoskeletons that can autonomously support human movement in unstructured, everyday environments by disengaging assistance when it is least likely to be appropriate.
\end{abstract}


\newcommand{\train}{$D_{\text{train}}^{\text{ID}}$}

\newcommand{\val}{$D_{\text{val}}^{\text{ID, OOD}}$}
\newcommand{\valID}{$D_{\text{val}}^{\text{ID}}$}
\newcommand{\valOOD}{$D_{\text{val}}^{\text{OOD}}$}

\newcommand{\testOffline}{$D_{\text{test (offline)}}^{\text{ID, OOD}}$}
\newcommand{\testOfflineID}{$D_{\text{test (offline)}}^{\text{ID}}$}
\newcommand{\testOfflineOOD}{$D_{\text{test (offline)}}^{\text{OOD}}$}

\newcommand{\testOnline}{$D_{\text{test (online)}}^{\text{ID, OOD}}$}
\newcommand{\testOnlineID}{$D_{\text{test (online)}}^{\text{ID}}$}
\newcommand{\testOnlineOOD}{$D_{\text{test (online)}}^{\text{OOD}}$}

\newcommand{\test}{$D_{\text{test (online)}}^{\text{ID, OOD}}$}
\newcommand{\testID}{$D_{\text{test (online)}}^{\text{ID}}$}
\newcommand{\testOOD}{$D_{\text{test (online)}}^{\text{OOD}}$}

\section{Introduction}

\IEEEPARstart{L}{ower} limb exoskeletons are used in a variety of applications to increase human mobility, from assistance to rehabilitation \cite{abery_role_2025, bartolo_exoskeletons_2021, baud_review_2021}. Recent results from human-in-the-loop optimization experiment have highlighted their potential to augment energy economy or user preferences during walking, running, and stair climbing, among many other tasks \cite{zhang2017human, slade2024human, ingraham2022role, ingraham2023leveraging, park2025human, ding2018human}. These results, however, can be contrasted with the continuing rarity of lower limb exoskeletons in the real world, particularly for populations who could most use assistance across a broad range of activities of daily living, but for whom safety is a necessity. For these potential users, the control laws that define when and how much assistance the exoskeleton provides must be predictable and robust.

Most exoskeleton controllers are designed and optimized for a few cyclic actions such as steady-state walking and jogging, climbing stairs, and other repetitive movements \cite{siviy_opportunities_2022}. This does not encompass the wide array of motions that people can perform in their daily lives. To address the limitations of steady-state controllers, data-driven controllers, especially deep learning-based controllers that handle temporal data, have shown promise in adjusting to various speeds and actions with adaptable torque control \cite{medrano2023real, shepherd_deep_2022, divekar2024versatile, belal2024deep, kang_real-time_2020, slade_personalizing_2022, strick2025real, hsu2024robustification, ronsse2011oscillator, zhang2024lower}. Deep learning-based controllers have shown promise in assistive and therapeutic settings \cite{shepherd_roadmap_2026, van_der_kooij_ai_2025}, and recently, deep learning biological torque controllers have represented a major step towards task-agnostic control \cite{molinaro_task-agnostic_2024}. However, these learned controllers have the major problem of faulty actuation based on out-of-distribution data; that is, if the model receives data that is faulty (e.g., dropped sensor) or outside of its training distribution, it will still make a prediction and the exoskeleton will act, without knowing whether the prediction was accurate. Critically, if the user is performing an action that the model was never trained on (whether gait phase, biological torque, or anything else), the prediction is very likely to be poor.

When exoskeletons actuate in the wrong direction or at the wrong time, the consequences can be disastrous for safety and trust in the exoskeleton. This may be most important at the ankle when the foot is suspended above the ground, as actuating during these times may cause tripping. If assistance is unexpectedly removed from 2-5\% of steps, user trust in the system has been shown to decrease \cite{wu2023impact}. Measures of exoskeleton trust are still being defined by the research community, but the harmful effect of failed assistance on trust, biomechanics, and safety can be assumed to be exacerbated in the presence of a wider variety of walking speeds, additional tasks, and transitions between task-specific controllers. There may also be more detriment to the exoskeleton mistakenly actuating instead of mistakenly turning off \cite{stirling_measuring_2024}.
\overviewFigure
A naive solution to the issue of unknown actions is to collect a training dataset consisting of all possible actions and label whether each task is assistable; however, due to the vastness of possible human actions, this is unlikely to be achievable. Another proposed solution is to develop machine learning controllers that can generalize well enough beyond the training set to provide useful assistance in new actions. There has been limited success in this endeavor. Phase-based controllers can extrapolate to different speeds of walking and different inclines, but have not been shown to generalize to new activities. The biological torque control approach taken by Molinaro, Scherpereel et al. demonstrates some ability to generalize to unseen activities \cite{molinaro_task-agnostic_2024}, largely maintaining the correct direction of torque, but accuracy dropped as low as $R^{2}$=0.24 for extremely dynamic tasks outside of its training set.

Without an all-encompassing training set or a perfectly generalizing controller, the safest thing for an exoskeleton to do when encountering an unknown movement or a movement the controller is not trained for may be to \textit{turn off}. To do this, however, the controller must be aware of whether the movement is unknown or novel to the controller. We propose running an uncertainty estimator -- in parallel with existing exoskeleton controllers -- to determine whether an action is known (in-distribution) or unknown (out-of-distribution), and ultimately to decide whether to activate the exoskeleton controller or to turn off (Fig.~\ref{fig:overview}). 

Contrary to the task classification strategies used in exoskeletons to date, our proposed uncertainty-aware control approach employs a one-class classifier; it is designed only using examples from the in-distribution class and must determine if the new samples are meaningfully different. Due to the high dimensionality of the input data, which consists of both sensor and time dimensions, deep learning-based methods may be advantageous. Across a variety of fields, including healthcare monitoring, manufacturing processes, and detecting fraud and fault, among others, deep learning-based uncertainty awareness (similarly novelty detection or anomaly detection) has been widely studied in recent years  \cite{schmidl_anomaly_2022, blazquez-garcia_review_2020}. For these purposes, deep learning is commonly used, especially for multivariate data \cite{iqbal_anomaly_2024, li_deep_2023, choi_deep_2021}. In estimating the uncertainty of neural networks specifically, state-of-the-art applications used methods like reinforcement learning and Bayesian belief networks \cite{lutjens_safe_2019, loquercio_general_2020}. This uncertainty awareness has been expanded to robot navigation and collision avoidance for reinforcement learning-based policies that can be used in real time \cite{kahn_uncertainty-aware_2017}. In addition to mathematical models, autoencoders, GANs, and ensembles of models have been proposed and found some success as architectures for anomaly detection \cite{an_variational_2015, mattia_survey_2021, iqbal_anomaly_2024, nawaz_ensemble_2024}.  Within biomechanics and assistive devices, Zhu et al. used deep learning-based anomaly detection to monitor the performance of wearable sensors in a prosthetic socket \cite{zhu_using_2023}. Hobbs et al. used a PCA-based method to identify outliers in synthetic gait data based on EMG sensors and kinematic data, but it is not a real-time method and is not generalizable to actions beyond walking \cite{hobbs_systematic_2022}.  Within the field of lower limb assistive exoskeletons specifically, deep learning-based controllers have been increasingly used to work towards task-agnostic control due to their ability to synthesize complex biomechanical sensor data into actionable control commands in real time, but to our knowledge, our method is the first deep learning uncertainty estimator to make real-time decisions about wearable robotic assistance.

The uncertainty estimator proposed in this study functions as a
high-level controller above whichever mid-level controller is in use.
Here, the mid-level controller for in-distribution actions is a walking
and jogging phase-based controller \cite{shetty_ankle_2025}, and the out-of-distribution action is disengagement, representing, to our knowledge, the first application
of uncertainty awareness for real-time safety-gating on an
exoskeleton. We use ``safety-gating" to describe this architecture: out-of-distribution detection is a statement about whether incoming sensor data resemble the training distribution, not a direct measure of whether the base controller is unsafe on that input. Disengaging is not a universally safe fallback (particularly for users who depend on assistance) and while we demonstrate detection and disengagement, we do not directly measure safety outcomes such as near-falls or adverse events; establishing these is left to future work. A controller that disengages on unfamiliar actions is nonetheless likely safer than one that applies the same assistance regardless of how well-suited it is to the action. 

The primary contributions of this work include 1) a foundational exploration of estimating uncertainty across tasks in human movement, 2) comparison of three deep learning-based methods and a classical baseline method with high potential in this application domain, and 3) the first real-time validation of an uncertainty-aware framework on an actuated lower limb exoskeleton in outdoor environments. Each uncertainty-aware model was trained on a 9-subject training set of exoskeleton sensor data (on board IMUs and ankle angle encoders) consisting of people walking and jogging on a treadmill (various inclines). The models were tuned on a 3-subject offline validation set, and tested on a new set of in- and out-of-distribution data. The best performing model was then tested online on an outdoor circuit with 3 new users. Our results demonstrate the feasibility of an uncertainty-aware exoskeleton controller for detecting and disengaging on unknown tasks. This approach offers a practical alternative to developing task-specific controllers for every possible movement—a currently intractable challenge that has hindered the real-world implementation of exoskeletons.

\sensorWindowsFigure

\section{Human Movement Data for Model Training and Development}
\label{sec:dataset_description}

Our approach is not tied to a particular exoskeleton, sensor suite, or task set: it requires only (i) a stream of onboard sensor data, (ii) a definition of the in-distribution set implied by the base controller's design, and (iii) a scalar uncertainty score that disengages assistance when it exceeds a threshold. We developed and evaluated the models entirely offline before online deployment. To preserve the principle that the uncertainty estimator sees only in-distribution data during training, and to avoid leakage between development stages, four datasets were used, each serving a distinct role: a training set \train, which is in-distribution by definition; an offline validation set \val, used for model development and  hyperparameter tuning; a held-out offline test set \testOffline, used only for final offline evaluation; and an online test set \testOnline, collected during real-time outdoor operation (Section V). The two development sets (\train and \val) are described below; the two test sets are described where they are used. All subjects provided informed consent, and this study was approved by the Northeastern Institutional Review Board.

The training dataset \train included 9 participants wearing ankle exoskeletons (EB60; Dephy Inc, Boxborough, MA) walking at various speeds and jogging at different inclines, previously presented in Shetty et al. \cite{shetty_ankle_2025}. Each task was performed under two conditions: actuated (using a simple Time-Based Estimator controller) and unactuated. Sensors include IMUs and ankle encoders on each boot (Fig. \ref{fig:sensor_windows}). Ankle velocity was also calculated from ankle angle via finite differencing and included as a channel.  Data were post-processed into overlapping time windows consisting of 175 samples, or approximately 1 s (windows were extracted at a stride of 10 samples ($\sim$0.057s), or every tenth window). It was previously found that the gait phase model accuracy did not substantially improve when using more windows \cite{shetty_ankle_2025}.

All actions were performed on a Force-Instrumented Treadmill (Bertec), which allowed us to calculate ground truth values for gait phase. Gait phase is used in our base controller: it is converted to a commanded exoskeleton torque via a spline, with spline parameters dependent on the predicted walking/jogging speed and ramp incline \cite{shetty_ankle_2025}. During the online validation (section V), this phase-based controller serves as a base controller for in-distribution movements.

The offline validation data \val\ was collected in the Northeastern motion capture laboratory with 3 subjects (1 subject from the training set, 2 new participants). It consists of a range of both in-distribution and out-of-distribution activities.
Activities included:
\begin{itemize}
    \item In-distribution: overground and treadmill walking and jogging (actuated and unactuated)
    \item Out-of-distribution: (stationary postures, chair-based movements, standing exercises, jumping, backward walking, etc).
\end{itemize}
In-distribution tasks were chosen based on the available data in \train, and due to the well-tuned exoskeleton controllers available for cyclic tasks like walking and jogging. The definition of in-distribution was applied conservatively in this study, for example, stairs were considered out-of-distribution despite ramps being included in the training dataset. 
\val\ was prepared in the same manner as \train. \val\ was not used in model training, though it was used to evaluate the models as we developed them.  \testOffline was collected similarly to \val on 3 new participants, with some new tasks added, including incline and decline walking, lying down, new sitting activities, standing poses, etc.  Table \ref{tab:novelty_overview} (see Appendix) contains a detailed breakdown of which tasks are included in each dataset.

\section{Uncertainty Estimation Methods}
\label{sec:model_architecture}
For uncertainty estimation, we primarily focused on deep learning techniques, since our data is high-dimensional, multivariate, and temporal. All deep learning models used Temporal Convolutional Networks (TCNs) in their first layers due to their effectiveness in previous exoskeleton research \cite{shetty_ankle_2025, molinaro_subject-independent_2022, kang_real-time_2021}.  TCNs use dilated causal convolutions that can be expressed as:
\begin{equation}
F(s) = (x *_d f)(s) = \sum_{i=0}^{k-1} f(i) \cdot x(s-d \cdot i),
\end{equation}
where $x$ is the input sequence, $f$ is the filter, $k$ is the filter size, and $d$ is the dilation factor.
Fig.~\ref{fig:architecture} illustrates the three deep learning-based uncertainty classifier architectures implemented in this study, all of which employed TCNs.

We implemented three deep architectures, a model ensemble
(Section~\ref{subsec:ensemble}), an autoencoder
(Section~\ref{subsec:autoencoder}), and a GAN (Section~\ref{subsec:gan}),  and compared them against a classical
one-class support vector machine (OCSVM) baseline
(Section~\ref{subsec:ocsvm}).
\subsection{Ensembles}
\label{subsec:ensemble}
Model ensembling is an established anomaly detection technique that leverages the differences between multiple similar models to detect anomalies in data  \cite{nawaz_ensemble_2024}. We hypothesized that if multiple TCN models were trained on in-distribution data (e.g., starting from different initializations or training on different subsets of the data), their predictions would agree when presented with new sensor data that is similar to the training set but diverge when encountering unfamiliar sensor data. Thus, increased variance between model outputs would indicate out-of-distribution actions. 

Our ensemble approach consists of 7 TCN models, each predicting the sine of gait phase for both the left and right leg. We found that including more than 7 models did not further improve performance. To avoid the discontinuity at heel strike when gait phase switches from 100 percent to 0 percent, we predict the sine of gait phase similar to Kang et al. \cite{kang_real-time_2020}. Each model $i$ in the ensemble had two heads: one head predicts the sine of the gait phase for the left leg, $l_i$, and the other head predicts that for the right leg, $r_i$. At each timestep, for input data $\mathbf{x}$, the uncertainty estimate $\Psi$ is computed as the average of the variance of predictions across the left, $l_1,\ldots, l_7$ and right legs, $r_1,\ldots, r_7$,
\begin{equation}
\Psi(\mathbf{x}) = (Var(l_1,\ldots, l_7) + Var(r_1,\ldots, r_7))/2.
\end{equation}
The hyperparameters of the TCN were taken from the gait phase model in previous work \cite{shetty_ankle_2025}. The models were trained with one subject withheld as the early stopping subject. This subject was rotated through the 9 subjects to find the average number of epochs before the models overfit. The final model was then trained for this number of epochs with all subjects included. 

The ensemble method as described so far has a disadvantage of requiring a ground truth label for its prediction variable--in this case gait phase, which already existed in our previously collected training data. Similar to other possible labels used in exoskeleton control, such as biological joint torque, gait phase can be costly to collect and process (and may require extra sensing). Furthermore, gait phase is not well-defined for non-cyclic tasks, meaning that in-distribution training data can only be cyclic actions like walking and running. This makes it less generalizable to future implementations that include a non-cyclic action in their in-distribution set. 

An alternate strategy could be to create a synthetic target that generalizes to a broader set of actions in the in-distribution test set. We explored several simple targets (e.g., means of channels, weighted polynomial combinations of channel means, etc). We found positive results for a synthetic target consisting of the sum of all correlations between all sensor channels across the window of input data. Thus, we similarly tested TCNs trained to predict these correlations (instead of sine of gait phase), and our results will distinguish between ``Ensemble Method (Gait Phase)'' and ``Ensemble Method (Synthetic Target).''
\subsection{Autoencoder}
\label{subsec:autoencoder}
To eliminate the dependence on labels, we explored autoencoder architectures, which are widely used for anomaly detection on time series data and require only unlabeled data ~\cite{schmidl_anomaly_2022, choi_deep_2021, zhou_contrastive_2022, asimakis_autoencoders_2024, thill_temporal_2021}. 
 Our autoencoder architecture employs a TCN-based encoder and decoder, and attempts to reconstruct the input data itself:

\begin{equation}
\mathbf{z} = f_{\text{enc}}(\mathbf{x}), \quad \hat{\mathbf{x}} = f_{\text{dec}}(\mathbf{z})
\end{equation}

Autoencoders create a bottleneck in an intermediate layer, forcing the network to learn a compressed representation of the training data. Typically, reconstruction error ($x - \hat{x}$) is used as the uncertainty score, since out-of-distribution data is less likely to reconstruct well. However, as described in the offline results (Section \ref{sec:offline-results}), reconstruction error did not work well as an uncertainty score for our application, likely because \valOOD\ contains a substantial amount of low-complexity stationary data, such as standing, which is particularly easy for the model to reconstruct.

Instead, we focus on the latent representation, which captures essential features of the training data. We hypothesized that in-distribution data will cluster in the latent space, with anomalous data falling outside these regions. Therefore, we used Local Outlier Factor (LOF) on the latent space. LOF is an unsupervised anomaly-detection technique that works well in low-dimensional problems; it evaluates the local density of each sample relative to its neighbors:  

\begin{equation}
\Psi(\mathbf{x}) = -\text{LOF}_k(\mathbf{z})
\end{equation}
The initial parameters chosen for the autoencoder were drawn from TCNs used in previous work, with the dominant hyperparameters associated with latent space size explored further in Section \ref{sec:offline-results} \cite{shetty_ankle_2025}. 
\architectureFigure
\subsection{Generative Adversarial Network (GAN)}
\label{subsec:gan}
The final architecture we explored is a Generative Adversarial Network (GAN), an adversarial approach consisting of a generator $G$ and a
discriminator $D$. Unlike the autoencoder, which uses reconstruction as
an intermediary, a GAN attempts to model the in-distribution data
distribution directly and classify inputs as in- or out-of-distribution
on a continuous scale. The generator learns to turn noise into outputs
that mimic in-distribution data, while the discriminator learns to
separate real in-distribution data from generated data; by modeling the
distribution directly, a GAN offers the possibility of capturing more
complex patterns while training only on in-distribution data.

Our generator uses a TCN-based decoder and our discriminator a TCN-based
encoder, trained with the standard adversarial loss:
\begin{equation}
\min_G \max_D \mathcal{L}_{\mathrm{GAN}}(G, D) =
\mathbb{E}_{\mathbf{x}}[\log D(\mathbf{x})] +
\mathbb{E}_{\mathbf{z}}[\log(1 - D(G(\mathbf{z})))]
\end{equation}
For our uncertainty measure, we use the discriminator output directly,
which lies between 0 and 1 and reflects its confidence that an input is
in-distribution:
\begin{equation}
\Psi(\mathbf{x}) = 1 - D(\mathbf{x})
\end{equation}

This approach can learn complex patterns but is known to be unstable in
training~\cite{goodfellow_generative_2014}, and it was the most difficult
of our architectures to tune: because the discriminator classifies rather
than generates, it tended to overpower the generator. To balance the two,
we weakened the discriminator using spectral normalization, fewer
discriminator updates, and a lower discriminator learning rate than the
generator (Table~\ref{tab:all_model_specs}). Even so, as reported in
Section \ref{sec:offline-results}, the GAN's sensitivity to these choices
limited its performance relative to the ensemble methods.

\subsection{One-Class SVM Baseline}
\label{subsec:ocsvm}

As a classical reference point, we include a one-class support vector
machine (OCSVM) with a radial basis function (RBF) kernel
\cite{scholkopf_estimating_2001}, a standard anomaly-detection method. Unlike
the deep architectures, an OCSVM has no mechanism to exploit the temporal or
cross-channel structure of a raw $175 \times 16$ window, and its RBF kernel
degrades as dimensionality grows and pairwise distances concentrate, which
also makes the kernel width difficult to set. A fixed front end therefore
first reduces each window to a compact, decorrelated representation: the
window is subsampled to every fifth timestep and flattened ($35$ timesteps
$\times$ $16$ channels $= 560$ features), amplitude-normalized by its
root-mean-square, and projected onto its first $20$ whitened principal
components ($58\%$ of training variance).

The OCSVM is then fit on this representation of the same in-distribution
training set $D_{\mathrm{train}}^{\mathrm{ID}}$ used for the deep models (all
$9$ subjects, all trials), with $\nu = 0.01$ and kernel width
$\gamma = 0.50$; its uncertainty score $\Psi(\mathbf{x})$ is the negative
signed distance to the decision boundary, so larger values indicate more
out-of-distribution inputs. Both stages are fit on
$D_{\mathrm{train}}^{\mathrm{ID}}$ only, and all subsequent processing is identical to that used for
the deep models, so the baseline differs from the deep architectures only in
the detector. 

\section{Offline Testing}

\subsection{Training}
\methodsFigure
All models were trained only on the in-distribution training set \train. This is crucial for our end objective: a framework future researchers can apply to their own in-distribution tasks without collecting out-of-distribution data, which can never be fully representative of human movement. Uncertainty-detection models are, however, harder to develop and tune without overfitting than typical exoskeleton controllers, since development requires an offline set containing both in- and out-of-distribution data, and that set must be long enough to hold out a subset of tasks for final testing. As described in Section~\ref{sec:dataset_description}, we therefore used \val for development and withheld \testOffline entirely (Table~\ref{tab:novelty_overview} gives the task composition of each set).

Concretely, models were trained on \train and run on \val, whose performance and J-statistic guided architecture changes and hyperparameter tuning. Once a model performed satisfactorily on \val, it was run on the withheld \testOffline for the offline results in Section~\ref{sec:offline-results}. The best-performing model was then run on \testOnline, 3 novel subjects on a continuous outdoor course (Section~\ref{sec:online-model-testing}). This process is shown in Fig.~\ref{fig:methods}.

In addition to hyperparameter tuning, the performance of uncertainty classification is highly sensitive to the uncertainty threshold, which controls the balance between the false positive and false negative rates. A threshold could be derived from \testOffline to optimally balance these false positive and negatives; however, we considered our findings to be more powerful if we fixed the threshold using only training data \train. In the best case scenario, this would enable other researchers to implement our methods without needing to collect out-of-distribution samples, and expect the method to work out-of-the-box. We set the threshold as the value that contained 99.5 percent of the uncertainty scores when using only \train, which we considered to represent a reasonable (0.5\%) fixed false positive rate. Notably, fixing the threshold in this manner may cause the threshold to be highly dependent on a few outlier movements in the training set. We note that this applies to the deployment threshold only: although the threshold itself is derived solely from \train, \val, which contains OOD data, was still used during model development to compare architectures and tune hyperparameters, as described above.

For all models, an 88 point ($\sim$0.5s) causal simple moving average filter was applied to the uncertainty scores. The vast majority of the frequency content of human movements is below $\sim$10 Hz \cite{winter2009biomechanics}; filtering the output removes spikes in the predictions that do not represent true transitions between in- and out-of-distribution tasks. However, increasing the filter strength comes with added delay, causing poor performance during task transitions. We assume our 0.5 s filter to add ~0.25 s delay during transitions.

\subsection{Evaluation}
Performance was assessed using standard metrics: accuracy, recall, precision, F1-score, and J-statistic \cite{youden_jstat_1950}. Youden's J-statistic (0-100\%) is defined as:
\begin{equation}
    \text{J-statistic} = \text{Recall} + \text{Specificity} - 1,
\end{equation}
where recall represents the proportion of actual out-of-distribution instances correctly identified, and specificity represents the proportion of in-distribution instances correctly identified. We used the J-statistic in addition to the other standard metrics because it is more robust to heavy class imbalances, which are common in anomaly detection tasks and in our test sets \cite{ruopp_youden_2008}. \testOffline consisted of 80.1\% out-of-distribution windows, while \testOnline consisted of 57.1\% out-of-distribution windows. In this study, out-of-distribution represents the positive class. Fig \ref{fig:offline_results} demonstrates the model performance on various task groups in \testOffline to further clarify the model's performance on this dataset. 
\offlineResultsTable
To assess the quality of probabilistic predictions, we also evaluated calibration using Expected Calibration Error (ECE) and Brier score. ECE measures the average absolute difference between predicted probabilities and actual frequencies across confidence bins \cite{guo_calibration_2017}, with lower values indicating better calibration. Brier score quantifies the mean squared difference between predicted probabilities and true outcomes \cite{gneiting_strictly_2007}, where a score of 0 represents perfect calibration and 0.25 is equivalent to always predicting 0.5. For models that output unbounded uncertainty scores (ensemble variance and LOF scores), we converted these to calibrated probabilities using sigmoid transformation \cite{john_platt_probabilistic_1999}: $p = 1/(1 + \exp(-s(\text{score} - \text{threshold})))$, where $s$ is a steepness parameter and the threshold centers the sigmoid at $p=0.5$. The steepness parameter was optimized using leave-one-subject-out cross-validation to minimize Brier score, ensuring no subject-level data leakage in calibration parameter selection. For the GAN discriminator, which already outputs probabilities via its sigmoid activation, we used the discriminator output directly without recalibration. We report calibration metrics using the pre-set training-based threshold (99.5th percentile of training uncertainty scores) in Table~\ref{tab:offline_results}. 

To compare the effect of threshold choice, we also calculated the ECE and Brier score using a threshold post-hoc optimized from \testOffline. Threshold choice showed varying impact on calibration across architectures: minimal effect for the gait phase ensemble ($<$6\% change in both metrics), substantial improvement for the synthetic target ensemble (39\% and 32\% reductions in Brier and ECE, respectively), moderate improvement for the LOF autoencoder (16\% and 5\% reductions), and no change for the GAN discriminator as expected from its direct probability outputs. To further evaluate the model robustness to threshold choice, we calculated the area under the ROC curve as well as the J-statistic of the model using both thresholds, reported in Fig.~\ref{fig:offline_results} \cite{bradley_use_1997}.

To evaluate our proposed uncertainty classification approaches, we conducted offline testing across all architectures using \testOffline. The models are run in the following steps:
\begin{enumerate}
    \item Pass a window (175 samples, or $\sim$1 second) of scaled data into the pre-trained model and get the uncertainty score for this window. 
    \item Filter the output with an 88 point ($\sim$0.5s) filter.
    \item Label the window as in or out-of-distribution based on whether the filtered uncertainty score is above or below the threshold.
    \item Mark the window as correct if the label of the window is the same as the ground truth label of the window.
    \item Calculate the overall accuracy, precision, recall, F1 score, J-statistic, ECE, and Brier score. 
\end{enumerate}

\subsection{Offline Results}
\label{sec:offline-results}

\offlineResultsFigure

The gait phase ensemble method demonstrated strong performance on \testOffline (J: 92.3, AUROC: 0.993). The gait phase ensemble sometimes misidentified stair ascent and descent (both cyclic tasks), especially on low incline stairs, as well as forward lunge, kicking, and shifting weight. 
Table~\ref{tab:offline_results} presents the performance metrics for each method, and Fig \ref{fig:offline_results} shows each model's performance by task.

The synthetic target ensemble model performed poorly compared to the gait phase ensemble model in our evaluation framework (J: 70.9). However, the in-distribution and out-of-distribution predictions did maintain good separability, obtaining a J-statistic of 89.5 when using a threshold post-hoc optimized from \testOffline, and the AUROC was 0.988. This method demonstrates a promising direction for future research to allow ensemble methods to move away from gait phase labels. 

The autoencoder is not reliant on a gait phase label, but performed worse than both ensemble methods (J: 58.5, AUROC: 0.909). We performed a brute-force sweep of latent space dimensions to better understand the effect of latent space size (Fig.~\ref{fig:autoencoder}). We hypothesized that for a small latent space, the model would poorly reconstruct both in- and out-of-distribution data, leading to poor separability. For large latent spaces, the autoencoder would learn to pass the input sequence through with minimal compression leading to easy reconstruction for both in- and out-of-distribution data. At an optimally sized latent space, we hypothesized that the autoencoder would compress input data enough that the model would be able to reconstruct in-distribution data well but not out-of-distribution data, leading to a separable uncertainty score of reconstruction error. However, after observing the performance of models across our latent space size sweep, the \valOOD\ reconstruction was not meaningfully separated from \valID\ at any latent space size, and \valID\ consistently reconstructed worse than \train\ (Fig.~\ref{fig:autoencoder}). We hypothesize that the good reconstruction of \valOOD\ is due to many of our out-of-distribution actions being stationary, i.e., sitting, standing, performing exercises in place, etc. This is easy for an autoencoder to reconstruct, even without prior data, as little to no movement can be approximated as a 0 for each channel. 

The latent representations for \valID\ and \valOOD\ were more separable, with in-distribution data exhibiting a ring-like structure often seen in cyclic data (Fig.~\ref{fig:autoencoder}). To take advantage of this better separation in the latent space than in the reconstruction error, we used Local Outlier Factor on the latent space as our uncertainty score, so a new window would be evaluated on how far its latent space representation fell outside of the distribution of the latent space of the training data. Using this method, we were able to obtain a model with a J-statistic of 84.9 on \val. However, these results were not generalizable to \testOffline, with the model obtaining a J-statistic of 58.5.
\AutoencoderFigure

The GAN-based approach is not reliant on a gait phase label, and demonstrated good performance in labeling many out-of-distribution actions correctly. However, it labeled jogging as out-of-distribution despite it being well-represented in \train, hurting its overall performance (J: 63.0, AUROC: 0.929). In addition, its performance was highly sensitive to the hyperparameters chosen; thus the GAN may have been more likely than the other architectures to be overfit to \val, and its performance may not be generalizable to other actions. This hypothesis is somewhat supported by the GAN's worst performing out-of-distribution tasks in \testOffline, many of which are tasks that were not included in \val. 

The gait-phase ensemble retained the best accuracy, precision, F1, calibration, and per-task consistency (Table~\ref{tab:offline_results}), was the least sensitive to hyperparameters, therefore we selected it for online implementation and further testing.
    
As noted in Section \ref{subsec:ocsvm}, an OCSVM's behavior depends heavily on its input representation and, for the RBF kernel, on the kernel width, so we swept nine window representations (varying subsampling stride, amplitude normalization, anti-aliased decimation, and channel-covariance descriptors) against ten detector settings. Performance was highly sensitive to these choices, most consequentially the RBF kernel width $\gamma$: on an identical representation, the default scikit-learn heuristic gave near-chance separation (AUROC 0.486), while a wider kernel ($\gamma$ = 10× scale) recovered strong ranking on \val (AUROC 0.988). Critically, this step required out-of-distribution data (\val) to rank configurations and to identify one whose threshold generalized. The baseline is therefore a strong competitor only when such validation data are available; without it, there is no reliable way to configure it, and its out-of-the-box behavior is poor.

With that caveat, the tuned baseline separated the classes well on the held-out offline set (AUROC 0.982, J = 85.1 at the fixed threshold; accuracy 89.1\%, F1 77.9\%). It outperformed the autoencoder and GAN while requiring no gait-phase label, and rivaled the synthetic-target ensemble in ranking ability (AUROC 0.982 vs 0.988) and exceeded it at the deployable threshold (J = 85.1 vs 70.9), though the synthetic target edged ahead once its threshold was post-hoc tuned (J = 89.5). It was, however, less consistent across tasks (misclassifying several out-of-distribution actions such as pivoting, outdoor stairs, and picking up objects) and less well calibrated than the gait-phase ensemble (ECE 0.15, Brier 0.09, vs 0.03 and 0.03).
\section{Online Model Testing }
\label{sec:online-model-testing}
Testing models online is critical to demonstrating control approaches are viable. Because the exoskeleton is actuated (or sometimes unactuated), and the user's biomechanics are altered by assistance, the incoming sensor readings depend on the controller output. This feedback loop typically causes degraded performance relative to offline testing, and can even cause instabilities if implementation is poor. The online test introduced three novel subjects, real-time closed-loop actuation, and several genuinely new tasks (e.g., curb navigation and suspending the feet off the ground), alongside task categories also represented in the offline sets (Table \ref{tab:novelty_overview}). It therefore evaluates generalization to new subjects and to real-time actuation, and partial generalization to unseen task categories, rather than to a fully held-out task set. 

\subsection{Online Implementation}
For online and real-time evaluation, our implementation architecture consisted of three components implemented similarly to previous work, with the addition of our uncertainty estimator running in parallel with the gait phase estimator \cite{shetty_ankle_2025}:

\begin{enumerate}
    \item \textbf{Dephy ExoBoot Hardware:} Ankle exoskeleton equipped with IMUs and encoders that provide raw sensor data and execute torque commands.
    
    \item \textbf{Jetson Orin Nano:} Executes the model inference of both Uncertainty Estimator (Gait Phase Ensemble Model) and Gait Phase predictor.
        
    \item \textbf{Raspberry Pi:} Receives sensor data from the Exoboots at 175 Hz, performs basic processing and communications, and sends torque commands. It serves as the intermediary between the Jetson and the exoskeleton hardware.                     
\end{enumerate}

The model developed and tested offline was converted from a pytorch model to a TRT for more efficient online performance on the Jetson. The end-to-end system operates at 105 Hz. For online deployment the uncertainty score was smoothed with an 88-sample ($\sim$0.5 s) causal \textit{median} filter rather than the simple moving average used offline. The median is more robust to the isolated score spikes, and shares the same $\sim$0.5 s window, and therefore the same $\sim$0.25 s half-window group delay assumed in Section IV-A.

\subsection{Online Experiment Methods}

\CourseMapFigure
\onlineOverallResultsTable
\onlineFigure

We designed a test route around Northeastern University campus (Fig.~\ref{fig:course_map}) that included various terrains and actions to validate the system's performance in real-world conditions:
\interlinepenalty=0
\widowpenalty=0
\clubpenalty=0
\begin{enumerate}
    \item \textbf{Stair station}: Ascending and descending stairs.
    \item \textbf{Ramps and suspension station}: Ascending and descending a ramp, and hanging with participant's feet off of the ground.
    \item \textbf{Object interaction station}: Picking up and throwing away an item in a trashcan that opens with a foot pedal, and picking up objects off the ground.
    \item \textbf{Sitting station}: Transitioning between sitting and standing, sitting on a bench, tapping feet while sitting,  and moving positions while sitting.
    \item \textbf{Obstacle course station}: Jumping and walking on and off of the curb, and skipping.
    \item \textbf{Plyometrics station}: Squats, lunges, toe raises, jumping jacks (subject 1 only).
    \item \textbf{Steady state walking}: Walking around the course at a brisk pace.
    \item \textbf{Jogging}: Jogging around the course at a slow pace.
\end{enumerate}

Three novel subjects performed this circuit with the ensemble of gait phase estimators providing real-time uncertainty estimates (Fig.~\ref{fig:onlineFigure}). When the uncertainty score was below the threshold (predicting in-distribution), the phase-based exoskeleton controller was active and providing torque. When out-of-distribution, the exoboot actuators unspooled, providing zero impedance. The model outputs and applied torque were recorded, as well as video footage of the participant traversing the obstacle course, which was time-synced to the model outputs for later evaluation. For subjects 2 and 3, jumping jacks were removed from the circuit due to hardware issues on high impact tasks, likely due to hardware age and extended use. 
To generate ground truth labels for evaluation, we employed three external observers (not on the research team) to mark timestamps in the video recordings where they would judge an activity to be in-distribution (\testID) or out-of-distribution (\testOOD). Labelers were instructed to label a timestamp as in-distribution if the subject was walking or jogging for two or more steps, and label everything else as out-of-distribution. The ground truth label (as a function of time) was then defined by majority voting from the labelers. The online evaluation metrics were calculated by comparing model classifications against these ground truth labels. Agreement between observers was substantial (Fleiss' $\kappa$ = 0.83 $\pm$ 0.05; pairwise percent agreement 0.92 $\pm$ 0.02) with all three observers unanimous for 88\% of the recording. 56\% of the remaining disagreement fell within 1 s of a transition boundary (the median observer-to-consensus boundary offset was 0.14 s).

\subsection{Online Results}

\onlineStationsResultsTable

The time-course of the uncertainty score and individual model outputs are shown in Fig.~\ref{fig:onlineFigure}. The overall F-1 score and J-statistics for the online model in the novel circuit were 89.3\%  and 74.4\%, a decrease from the performance on the offline dataset where the F-1 score was 90.3\% and the J-statistic was 92.3\% (Table~\ref{tab:ensemble_transitions_overall}). The overall AUROC for the online testing was 0.943 $\pm$ 0.02, with per-subject values of 0.922, 0.939, and 0.967. This represents a slight decrease from the offline AUROC of 0.993. All sections had accuracies above 85\% except for the stair section which had an accuracy of 71.1\%  (Table~\ref{tab:ensemble_results_by_station}. Note that we report model performance by action group as accuracy, since each group is either fully in- or fully out-of-distribution). With the stair section removed, the model F-1 score and J-statistic improved to 92.5\% and 81.0\%, respectively. Assistance was withdrawn 0.33 ± 0.28 s after a true out-of-distribution transition, dominated by the 0.50 s median filter. A causal median follows a sustained change once more than half its window has crossed, resulting in a delay of half the window, 0.25 s by construction. The measured delay is slightly under the 0.25 s half-window (0.20 s) because the sliding input window already carries post-transition data at the transition, priming the filter. Torque was gated in the same control cycle ($<$6 ms) and the belt reached slack (so no torque could reach the ankle) 0.12 s later, so 5 of 11 transitions received 0.29 ± 0.18 s of mistimed assistance (peak 17 ± 7 Nm, 3.7 ± 3.0 Nm·s).

After observing that the first subject's performance on the stairs was worse than the rest of the sections, we collected online data of the remaining subjects ascending and descending a separate set of stairs. The stairs on the outdoor course had a lower incline than standard stairs, with a rise of 15 cm and a run of 41 cm. The separate set of stairs we collected fell within the standard dimensions of indoor stairs, with a rise of 18 cm and a run of 28.5 cm. The model's average online accuracy on these standard stairs was 88.9\%. Similar results can be seen in the model performance offline, as shown in Fig.~\ref{fig:offline_results} where the model misclassified the low incline stairs much more than the standard stairs. 

To further characterize the model's online performance, we defined transition regions as the reviewer-labeled timestamps for transitions between in/out-of-distribution $\pm$ 0.5s. These regions do not have an easy definition of in- vs out-of-distribution. We observed that the transition regions performed worse than the non-transition regions; removing the transitions increased the F-1 score and J-statistic to 90.8\% and 77.0\% including stairs, and 94.2\% and 84.7\% without stairs.

\section{Discussion}
This study demonstrates a novel uncertainty estimator for use in lower limb exoskeletons, allowing differentiation between in-distribution and out-of-distribution actions. Our test of the uncertainty estimator running in parallel with an ML-based gait phase estimator demonstrates how the uncertainty estimator can be implemented online in tandem with other actuation control algorithms that are designed for specific actions, but fail for others. The uncertainty estimator is shown to work online with actuation on novel subjects in a diverse outdoor environment on a multitude of different actions including but not limited to plyometrics, sitting, walking on ramps, interacting with objects, and walking on and off multi-level surfaces like curbs. The model achieved an F1 score of 94.2\% and a J-stat of 84.7\% (not counting transitions and stairs) on novel subjects.

Because the online test set introduced both novel participants and novel tasks with no matched offline counterpart, we do not interpret the offline-to-online change as a like-for-like comparison of performance on identical conditions. Rather, the consistency of the uncertainty estimator across two datasets with substantially different task compositions is itself the relevant result: it indicates that the estimator generalizes to unseen out-of-distribution tasks and to real-time actuation without a large loss in discriminative performance, rather than having overfit to the specific tasks in the offline set.

The largest decreases in performance came from the stair incline and decline portion of the course.  As demonstrated in Fig \ref{fig:offline_results} and the online performance on the separate set of standard stairs, the model performed much better on standard stairs both online and offline than it did on the low incline stairs. The flatter, wider steps that were a part of our course likely provided less of a distinction from ramps, causing the model to falsely label much of the stair ascent and descent as in-distribution and hurting the overall accuracy of the model. We suggest including both stairs and ramps with similar inclines into the same grouping in future implementations.

The offline test set did not include any transitions between in- and out-of-distribution actions, so some decrease in performance can be attributed to the addition of these transitions. Transitions can be difficult or impossible to define categorically, as they may be somewhere between in and out-of-distribution and there is no definitive moment of transition. In addition, the TCN has a window of 1 second of previous data that it uses to make its prediction for the current sample, so when a user has just transitioned to an out-of-distribution action, some of that window is still from the time when they were conducting an in-distribution action (and vice versa). Future implementations of this algorithm should include a gradual and smooth scaling of torque when transitioning between in-distribution and out-of-distribution movements. 

Even with some shortcomings in our current uncertainty estimator, it adds a safety-oriented fallback to the base controller, since the actions that are most distinct from the training data (e.g., jumping) have the fewest false positives (Fig. \ref{fig:offline_results}). For tasks that are clearly out-of-distribution, like standing still or jumping in the air, the uncertainty estimator performs very well. This can have the effect of preventing actuation at especially consequential moments like when the foot is off the ground, as demonstrated during the jumping tasks and during a task where the participant suspended their feet off the ground by leaning on a railing. Moreover, falsely actuating on tasks that are out of distribution but biomechanically similar to those in the training dataset is less likely to be safety critical, as the exoskeleton controller may still provide useful assistance. This asymmetry matters because novelty and danger are correlated but not identical: some out-of-distribution actions may still be handled acceptably by the base controller, so a missed detection is not necessarily harmful, while an in-distribution input could in principle be handled poorly. The safety benefit of our framework therefore comes not from a one-to-one mapping between novelty and danger, but from the fact that the most biomechanically distinct actions (where mistimed assistance is most consequential) are also the ones the estimator flags most reliably. Future work should improve uncertainty estimator performance to better handle out-of-distribution tasks that are close to in-distribution tasks.
    
The model was only trained on \train, which was collected with IMUs, encoders, and force plates (only needed for the ensemble of gait phase estimators).  Our offline validation set \val was used for model development and hyperparameter tuning, with the ensemble of gait phase estimators being the only model to achieve high enough performance on \testOffline to try online. The online implementation only requires the onboard exoskeleton IMUs and encoders, allowing it to be implemented in any location or environment. Future work could see attempting this same method on other exoskeletons to see how much or little this method needs to be changed to achieve similar results with other datasets. In addition,  for the ensemble of gait phase TCNs, the validation set was only used for evaluating the model before testing it offline and online, not in the construction of the model or its parameters (that is, it worked well ``out of the box"). If others were to implement a similar method for their own exoskeletons using cyclic data, a validation set may not be needed.

Our approach allows for wider use of existing exoskeleton controllers in different environments and on different actions. Many exoskeleton controllers have been designed specifically for a discrete set of tasks. Researchers have largely ignored how their controllers will perform outside of these tasks; for instance, time-based estimator and even gait-phase estimators are likely to provide unhelpful ``assistance'' when the user is doing non-cyclic tasks, or even cyclic tasks that are far from the intended use case such as skipping or walking backwards. Similarly, controllers based on neuromechanical models often rely on the user performing tasks where reflexes are active, or ignore antagonist muscles that may co-contract during highly dynamic movements \cite{firouzi2025biomechanical, durandau2022neuromechanical}. Our concept of a high-level ``actuate'' vs. ``do-not-actuate'' classifier, designed via uncertainty estimation, can be layered on top of these controllers, with the goal of enabling the devices to be used more safely outside of their designed task set. We anticipate that this will open up possibilities for many researchers to safely take their exoskeletons out of the lab and facilitate real-world implementation. 

Future directions include pursuing a version of this estimator that is not reliant on gait phase, as this limits the in-distribution action set to those with a definable gait phase.  The ensemble of models trained to predict the synthetic target can generalize to different training sets, and in the semi-supervised case with the tuned threshold, the performance was nearly equivalent; it is possible that future work in designing a synthetic target and/or defining the threshold could match the gait phase ensemble's performance. The autoencoder and GAN approaches were difficult to tune, and the GAN was particularly sensitive to the chosen hyperparameters. Future work can continue to refine these approaches.

To achieve the longer term goal of utilizing this approach on users that require mobility assistance, training datasets must be expanded to include patient populations. Pathological gait, due to its differences from able-bodied gait, may register as out-of-distribution, and future work should be done to make uncertainty estimators that can classify these differing gaits as in-distribution, whether through population-specific calibrations or more expansive general training sets. The uncertainty awareness framework may also be expanded by including computer vision, which may provide additional context about the wearer's external conditions. While there are many potential avenues of continued development and refinement, this study overall represents a promising first step into the use of real-time uncertainty estimators for exoskeleton control in unknown situations, a step towards the safe use of deep learning based task-agnostic control strategies.

\section{Conclusion}
This study successfully developed and validated an uncertainty-aware control framework enabling ankle exoskeletons to operate beyond controlled laboratory settings. Our ensemble of gait phase estimators achieved robust real-time uncertainty estimation, switching between the specialized controller and a disengaged fallback across diverse real-world activities, though performance degraded during transitions and on tasks biomechanically close to the training distribution. This framework represents a substantial advancement toward practically implementable exoskeletons that can provide context-aware assistance across everyday environments while reducing inappropriate actuation during unanticipated movements.



\bibliographystyle{IEEEtran}
\bibliography{references}

\section{Appendix}
\label{sec:appendix}
\noveltyTable
\allmodelspecstableonecol

\vfill

\end{document}